\documentclass[letterpaper]{article} 
\usepackage{aaai2026}   
\usepackage{times}  
\usepackage{helvet}  
\usepackage{courier}  
\usepackage[hyphens]{url}  
\usepackage{graphicx} 
\usepackage{natbib}  
\usepackage{caption} 
\usepackage{amssymb}
\usepackage{algorithm}
\usepackage{algorithmic}

\usepackage{newfloat}
\usepackage{listings}
\DeclareCaptionStyle{ruled}{labelfont=normalfont,labelsep=colon,strut=off} 
\floatstyle{ruled}
\newfloat{listing}{tb}{lst}{}
\floatname{listing}{Listing}
\usepackage{booktabs}
\usepackage{xcolor}
\usepackage{color, colortbl}
\definecolor{Gray}{gray}{0.9}
\usepackage{amsmath}
\usepackage{enumitem}
\usepackage{helvet}  
\usepackage{courier}  
\usepackage{subfigure}
\usepackage{amsfonts}       
\usepackage{multirow}

\newcommand{\tabref}[1]{Table~\ref{#1}}

\newcommand{\eqnref}[1]{Eq.~(\ref{#1})}

\newtheorem{theorem}{Theorem}

\title{BayesAgent: Bayesian Agentic Reasoning Under Uncertainty via \\Verbalized Probabilistic Graphical Modeling}
\author {
    Hengguan Huang\equalcontrib\textsuperscript{\rm 1,2}\thanks{Correspondence to Hengguan Huang.},
    Xing Shen\equalcontrib\textsuperscript{\rm 3},
    Guang-Yuan Hao\textsuperscript{\rm 4},
    Songtao Wang\textsuperscript{\rm 5},\\
    Lingfa Meng\textsuperscript{\rm 1},
    Dianbo Liu\textsuperscript{\rm 6},
    David Alejandro Duchene\textsuperscript{\rm 1},
    Hao Wang\textsuperscript{\rm 7},
    Samir Bhatt\textsuperscript{\rm 1,2}
}
\affiliations {
    \textsuperscript{\rm 1}University of Copenhagen\\
    \textsuperscript{\rm 2}Imperial College London\\
    \textsuperscript{\rm 3}McGill University\\
    \textsuperscript{\rm 4}Cornell University\\
    \textsuperscript{\rm 5}University of Alberta\\
    \textsuperscript{\rm 6}National University of Singapore\\
    \textsuperscript{\rm 7}Rutgers University\\

    hengguan.huang@sund.ku.dk, xing.shen@mail.mcgill.ca, hw488@cs.rutgers.edu, samir.bhatt@sund.ku.dk
}

\begin{document}

\maketitle

\begin{abstract}
Human cognition excels at transcending sensory input and forming latent representations that structure our understanding of the world. While Large Language Model (LLM) agents demonstrate emergent reasoning and decision-making abilities, they lack a principled framework for capturing latent structures and modeling uncertainty. In this work, we explore for the first time how to bridge LLM agents with probabilistic graphical models (PGMs) to address agentic reasoning under uncertainty. To this end, we introduce Verbalized Probabilistic Graphical Modeling (vPGM), a Bayesian agentic framework that (i) guides LLM agents in following key principles of PGMs through natural language and (ii) refines the resulting posterior distributions via numerical Bayesian inference. Unlike many traditional probabilistic methods requiring substantial domain expertise, vPGM bypasses expert‐driven model design, making it well‐suited for scenarios with limited assumptions. We evaluated our model on several agentic reasoning tasks, both close-ended and open-ended. Our results indicate that the model effectively enhances confidence calibration and text generation quality.
\end{abstract}

\begin{links}
    \link{Code and Appendix}{https://github.com/xingbpshen/agentic-reasoning-vpgm}
\end{links}

\section{Introduction}
In addressing complex reasoning problems, such as solving challenging science questions, the human brain is thought to have the capability to go beyond mere sensory input, potentially forming insights into latent patterns of the world. This ability suggests that humans might have a sophisticated skill to interpret the underlying structures and uncertainties \cite{tenenbaum2011grow}, although the exact mechanisms remain the subject of ongoing research and debate. As of now, such depth of understanding demonstrated by humans has not been fully achieved in artificial intelligence (AI) systems \cite{lake2017building,bender2020climbing, zheng2021does,sumers2023cognitive}. 

While large language models (LLMs) have demonstrated impressive capabilities in processing and generating human language \cite{devlin2018bert,brown2020language, achiam2023gpt}, their performance is often constrained by the scope of their training data. These models, built primarily on vast corpora of text, excel at generating responses that are syntactically coherent and contextually relevant. Recent advances such as chain-of-thought (CoT) prompting \cite{wei2022chain} and the emergence of agentic paradigms \cite{yao2023react,schick2023toolformer}  have extended their capabilities toward interactive and compositional agentic reasoning. However, when operating as autonomous agents in uncertain or partially observable environments, where implicit knowledge and the ability to integrate and reason over undisclosed information from multiple sources become essential, skills that humans typically employ in complex reasoning, LLM agents often struggle. This limitation arises not only from their dependence on surface-level linguistic correlations but also from the absence of a principled Bayesian framework to capture latent structures and model uncertainty.

In this work, we explore for the first time how to bridge LLM agents with probabilistic graphical models (PGMs) to address agentic reasoning under uncertainty. To this end, we introduce Verbalized Probabilistic Graphical Modeling (vPGM), a  Bayesian agentic framework that combines the strengths of LLM agentic reasoning with explicit numerical Bayesian inference. Unlike traditional Bayesian inference frameworks \cite{l2008bayesian, bielza2014bayesian, wang2020survey, abdullah2022review}, which typically require substantial domain expertise, vPGM bypasses expert-driven model design, making it well-suited for scenarios with limited assumptions. Specifically, Bayesian structure learning methods \cite{kitson2023survey} facilitate the discovery of Bayesian networks, they often require expert domain knowledge for manual validation of statistical dependencies or rely on computationally expensive scoring functions to assess the graphical model’s goodness of fit to the data. Our approach leverages the knowledge and reasoning capabilities of LLMs by guiding them to simulate Bayesian reasoning princples, while augmenting uncertainty quantification through a learnable Bayesian surrogate, thus significantly reducing the reliance on expert input.

Concretely, our method consists of three initial stages: 
(1) Graphical Structure Discovery, in which the
LLM is prompted to identify latent variables and their probabilistic dependencies; (2) Prompting-Based Inference, where LLMs are guided to infer verbalized posterior distributions of each latent variable given new input data; and (3) Predictions under Uncertainty, where confidence in the final predictions is achieved by computing the expected value of the conditional predictive distribution over the inferred latent variables. Furthermore, to fully leverage the multiple response samples generated by LLMs within the vPGM framework and enhance uncertainty quantification, we extend vPGM with \emph{numerical} Bayesian inference techniques that infer posterior distributions over predictions and augment confidence calibration through a theoretically guaranteed differentiable calibration loss function.

We evaluate our method on several agentic reasoning tasks, designed in both close-ended and open-ended answering formats. The experiments demonstrate improvements in confidence calibration and the quality of generated responses, highlighting the efficacy of vPGM in enhancing probabilistic reasoning capabilities of LLM agents.

\begin{figure*}[h]
    \centering
    \resizebox{1\textwidth}{!}{
        \includegraphics{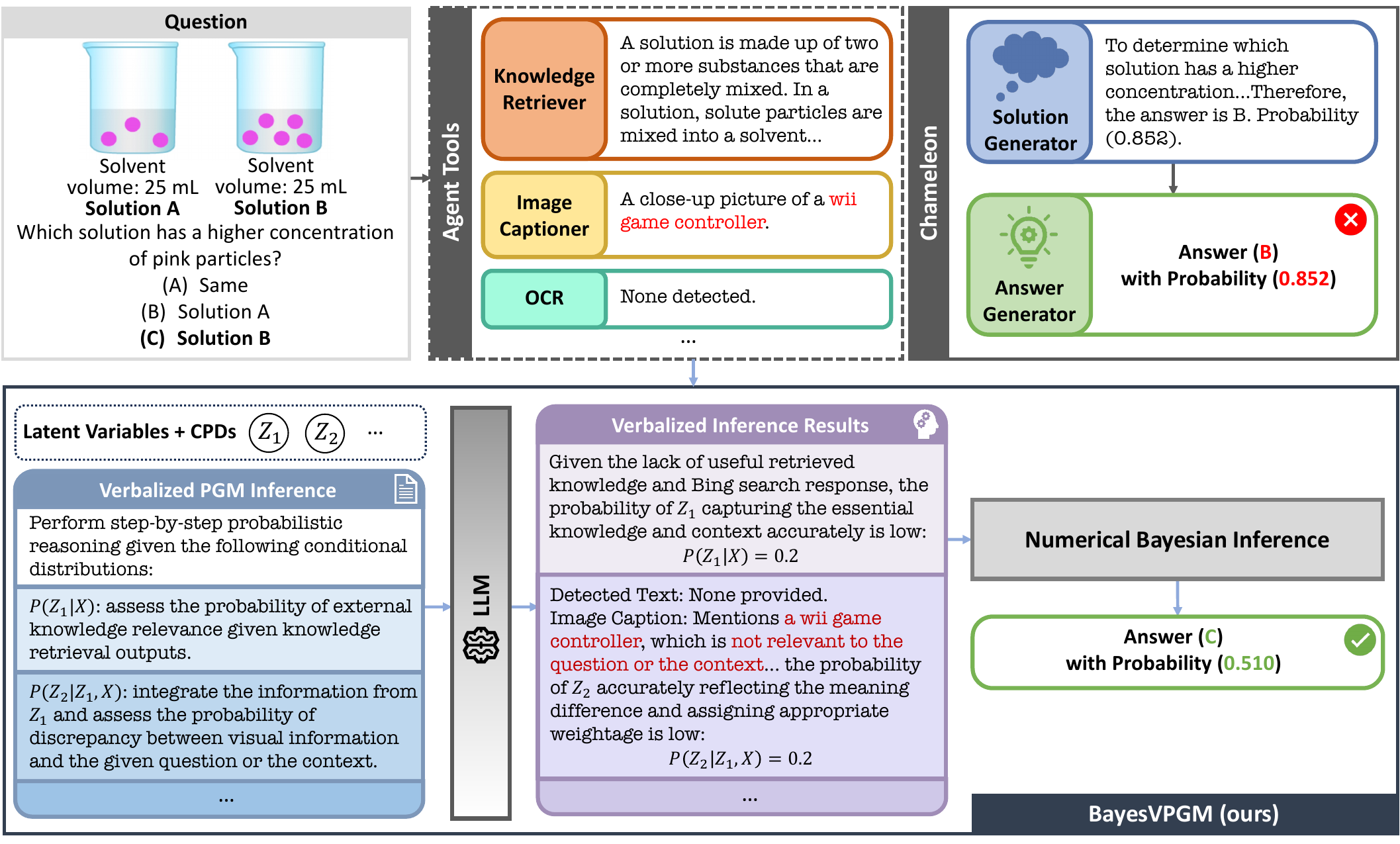}
    }
    \caption{Example of inference using the BayesVPGM. The Chameleon framework erroneously assigns high confidence to the answer despite its LLM agents capturing irrelevant information. Conversely, our BayesVPGM accurately identifies this discrepancy and assigns low confidence. Here, we show a simplified inference prompt. See Appendix for detailed examples.}
    \label{fig:overview_inference}
\end{figure*}

\section{Related Work}
\label{bg}
Research on large language models (LLMs) has recently transitioned from static prompting toward LLM agents or agentic systems capable of agentic reasoning, tool use, and interactive decision-making. We discuss both threads respectively, highlighting their limitations and how our proposed vPGM addresses a key missing component: probabilistic latent-variable reasoning and uncertainty calibration for agentic reasoning tasks.

\paragraph{LLM Prompting}
Prompting methods in LLMs form a long-standing research line centered on training-free model responses steering. Early approaches include in-context learning \cite{brown2020language}, where models are conditioned on task-specific demonstrations, and instruction prompting \cite{wang2022selfinstruct,ouyang2022training}, which embeds explicit task instructions directly into natural-language prompts. A major development is Chain-of-Thought (CoT) prompting \cite{wei2022chain}, which elicits intermediate reasoning steps to enhance complex reasoning. Subsequent variants extend CoT to more flexible or automated settings: zero-shot CoT \cite{kojima2022large}, automatic rationale generation (Auto-CoT) \cite{zhang2022automatic,shum2023automatic,yao2024tree}, self-consistency decoding \cite{wang2022self}, and chain-of-continuous-thought \cite{hao2024training}, which embeds reasoning trajectories in a latent space. Additionally, \cite{xiong2023llms} built upon the consistency-based method and conducted an empirical study on confidence elicitation for LLMs. In contrast, our proposed vPGM tackles the confidence elicitation problem from the perspective of Bayesian inference, which follows the principles of a more theoretically grounded Bayesian inference framework, PGM. 

\paragraph{LLM Agents and Agentic Systems}
Building on these prompting advances, LLM prompting has evolved into LLM agents, which interleave reasoning with actions, tool use, and interaction with external environments. ReAct \cite{yao2023react} combines natural-language reasoning with tool calls and environment feedback; Toolformer \cite{schick2023toolformer} uses self-supervised signals to teach LLMs when and how to invoke tools, and ADAS \cite{wang2025adas} automates the design of agentic system architectures. These systems mark a shift from passive text generation to interactive, tool-augmented behavior. However, existing agentic approaches typically lack a principled probabilistic framework: they do not explicitly model latent variables, quantify uncertainty, or perform Bayesian belief updating, which limits their applicability in settings that require calibrated agentic reasoning under uncertainty.

\paragraph{Concurrent Work}
Several concurrent works explore the use of LLMs for probabilistic or causal modeling, but they are largely orthogonal to our contribution. Recent causal-discovery studies \cite{wan2025llmcausaldiscovery,gpt4_causal_ml} focus on learning causal relationships and counterfactuals, whereas vPGM targets non-causal probabilistic latent-variable reasoning and uncertainty calibration for multi-source agentic tasks.  BIRD \cite{bird2025} introduces a Bayesian inference wrapper for LLMs, yet it is restricted to binary decision-making and is therefore not directly applicable to our multi-class and open-ended outputs. In contrast, vPGM provides a unified Bayesian framework for latent-variable reasoning and calibrated uncertainty within \textbf{LLM agents}.

\footnotetext{Although we set $n \le 4$ in this example, the LLM may generate the maximum number of variables. To reduce redundancies, we can add additional constraints to encourage a more compact representation.}

\section{Our Method: Verbalized Probabilistic Graphical Modeling (vPGM)}

Verbalized Probabilistic Graphical Modeling (vPGM) is a \emph{ Bayesian Agentic Reasoning} approach that leverages Large Language Models (LLM) agents to simulate key principles of Probabilistic Graphical Models (PGMs) in natural language. Unlike many existing probabilistic methods that  demand extensive domain knowledge and specialized training, vPGM bypasses the need for expert-based model design, making it suitable for handling complex reasoning tasks where domain assumptions are limited or data are scarce.

\subsection{Overview of vPGM}
\begin{figure}[h]
    \centering
    \resizebox{1\linewidth}{!}{
        \includegraphics{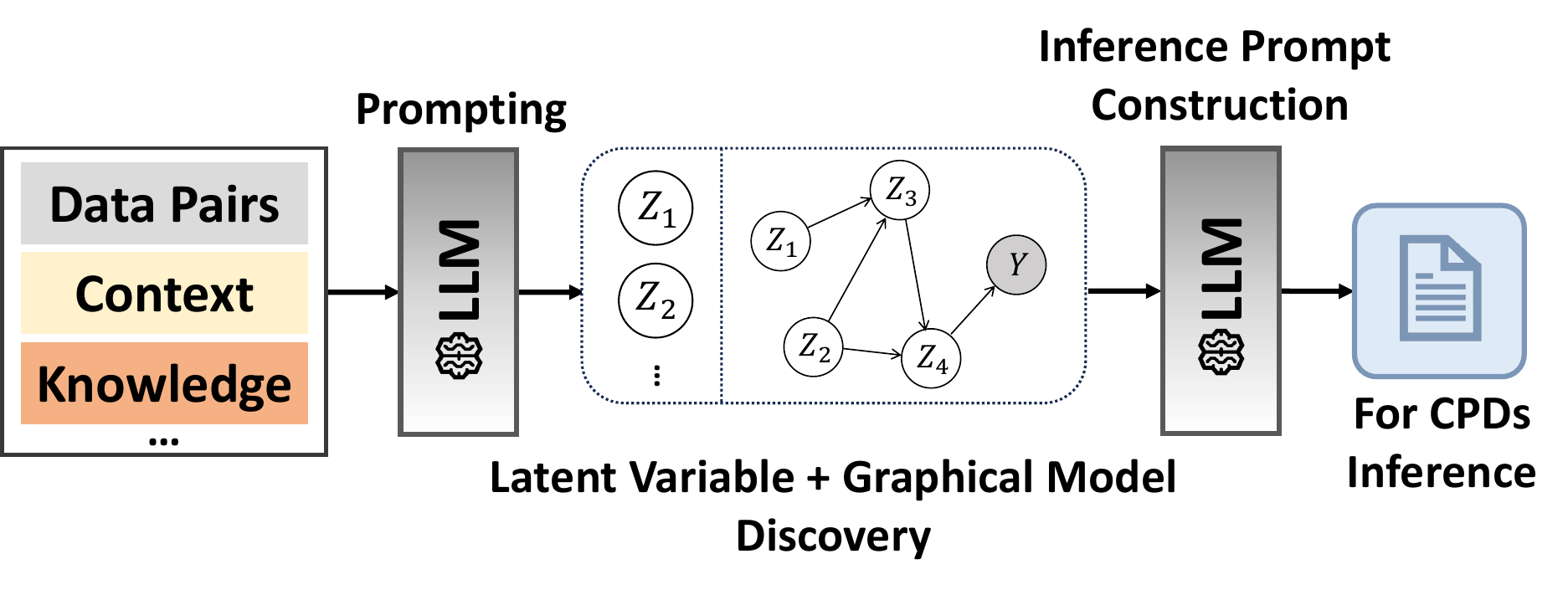}
    }
    \caption{Overview of the vPGM's learning framework. CPDs represent conditional probability distributions. We omit the observed variable $\mathbf{X}$ for clarity.}
    \label{fig:overview_learning}
\end{figure}

From an application standpoint, vPGM can be embedded into a range of complex reasoning systems, such as agentic reasoning tasks (see Figure~\ref{fig:overview_inference}). Our approach factorizes the overall reasoning process into three core steps: \textbf{(1) Graphical Structure Discovery}, in which the LLM is prompted to identify latent variables and their probabilistic dependencies (see Figure~\ref{fig:overview_learning}); \textbf{(2) Prompting-Based Inference}, where LLMs are guided to infer verbalized posterior distributions of each latent variable given new input data; and \textbf{(3) Predictions under Uncertainty}, where confidence in the final predictions is achieved by computing the expected value of the conditional predictive distribution over the inferred latent variables.

\subsection{Graphical Structure Discovery}
\label{sec:pgm_structure_discovery}
Our method begins by formulating a specialized prompt 
(see the appendix for the prompt)
to uncover latent variables for compositional reasoning. The prompt comprises several key elements: (1) General Task Description, a concise statement of the reasoning objective; (2) Input-Output Data Pairs, which illustrate representative data samples; (3) Contextual Information, providing any essential background or domain insights; and (4) Prior Knowledge and Constraints, specifying constraints such as the maximum number of latent variables and predefined dependencies among them.

After identifying a set of latent variables $\mathbf{Z} = \{Z_1, Z_2, \ldots, Z_n\}$ 
(see the appendix for examples of latent variables)
, we further prompt LLMs to determine how each latent variable depends on the others. An example set of dependencies obtained from the LLM is: $\{\mathbf{X}\rightarrow Z_1, \mathbf{X}\rightarrow Z_2, \mathbf{X}\rightarrow Z_3, \mathbf{X}\rightarrow Z_4, Z_1\rightarrow Z_3, Z_2\rightarrow Z_3, Z_2\rightarrow Z_4, Z_3\rightarrow Z_4, Z_4\rightarrow \mathbf{Y}\}$, where each relationship $a \rightarrow b$ indicates that $b$ is conditionally dependent on $a$. Like traditional PGMs, our verbalized PGM (vPGM) encodes these dependencies as conditional probability distributions $P\bigl(Z_i \mid \mathrm{Pa}(Z_i)\bigr)$. However, instead of relying on explicit distributional forms, vPGM uses natural language descriptions (see Appendix for detailed examples) to specify each conditional relationship, reducing the need for extensive domain expertise or parameter estimation.

\subsection{Prompting-Based Bayesian Inference}
\label{sec:bayesian_inference_as_prompt}
Traditionally, Bayesian inference focuses on inferring posterior distributions over model parameters given a probabilistic model and new observations. In the context of LLMs, however, it is reformulated as generating prompts that simulate posterior inference under the vPGM framework, leveraging its discovered structure and new observations. This approach leverages the advanced reasoning capabilities of LLMs to produce instructions enabling them to simulate Bayesian inference principles. An example prompt is: \emph{``Generate the prompt that guides LLMs through step-by-step probabilistic reasoning based on the provided task description, discovered PGM, and testing data...''}

\subsection{Prediction Under Uncertainty}
\label{sec:prediction_uncertainty}
Agentic reasoning tasks often involve significant uncertainty. For instance, an LLM agent (e.g., an image captioner) may produce noisy outputs, introducing aleatoric uncertainty. Under the vPGM framework, this variability is captured by the verbalized posterior distributions of latent variables. After constructing the verbalized posterior \(P(\mathbf{Z} \mid \mathbf{X})\) via prompting-based Bayesian inference, we quantify confidence in the final predictions by taking the expected value of \(P(\mathbf{Y} \mid \mathbf{Z})\) over \(\mathbf{Z}\):
\begin{equation}
    \mathbb{E}_{P(\mathbf{Z} \mid \mathbf{X})} \bigl[P(\mathbf{Y} \mid \mathbf{Z})\bigr]
    \;\approx\;
    \sum_{\mathbf{Z}} P(\mathbf{Y} \mid \mathbf{Z}) \, P(\mathbf{Z} \mid \mathbf{X}),
\end{equation}
where \(\mathbf{X}\) denotes observed inputs, and \(\mathbf{Z}\) is sampled by querying LLM using vPGM's Bayesian inference prompt. In practice, both \(P(\mathbf{Z} \mid \mathbf{X})\) and \(P(\mathbf{Y} \mid \mathbf{Z})\) are simulated within a single prompt (see detailed examples in the Appendix). Consequently, the expected posterior probabilities can be approximated by averaging the numerical values of \(P(\mathbf{Y} \mid \mathbf{Z})\) generated by the LLM during these inference steps.

\section{Bayesian-Enhanced vPGM: BayesVPGM}
\label{sec:bayesian_conf_cal}
When repeatedly querying a Large Language Model (LLM) under the vPGM framework, we obtain multiple samples of responses, i.e., categorical predictions and their numerical probabilities. A natural question is how to leverage these data to better capture the underlying uncertainty in the LLM's predictions.  To do this, we propose to infer such a posterior distribution, denoted \(q(\mathbf{y}\mid \tilde{\mathbf{x}})\), where \(\tilde{\mathbf{x}}\) denotes categorical predictions.

\subsection{Posterior Inference Under a Dirichlet Prior}
We specify the form of the posterior $q(\mathbf{y}\mid \tilde{\mathbf{x}})
  \;=\;
  \mathrm{Cat}(\boldsymbol{\pi}),$
where \(\boldsymbol{\pi} = (\pi_{1},\dots,\pi_{K})\) lies in the probability simplex over \(K\) categories. To incorporate prior beliefs, we place a Dirichlet prior on \(\boldsymbol{\pi}\):
\(
  \boldsymbol{\pi}
  \;\sim\;
  \mathrm{Dirichlet}(\alpha_{1},\dots,\alpha_{K}),
\)
with \(\alpha_{k} = \lambda\,p(y=k\mid \mathbf{Z})\) for some hyperparameter \(\lambda>0\), reflecting the vPGM’s initial belief in category~\(k\).  

Next, suppose we query the LLM under the vPGM framework for \(n\) times, obtaining labels \(\{y_{1},\dots,y_{n}\}\).  For each category \(k\), let \(n_{k}\) be the number of labels that fall into that category.  Assuming these labels are drawn i.i.d.\ from 
\(\mathrm{Cat}(\boldsymbol{\pi})\), 
the likelihood is $P\bigl(\{y_{i}\}\mid\boldsymbol{\pi}\bigr)\;=\;\prod_{k=1}^K \pi_{k}^{\,n_{k}}.$
By Bayes’ rule, the posterior distribution is then 
\[
q(\mathbf{y}\mid \tilde{\mathbf{x}})
\;\propto\;
\Bigl(\prod_{k=1}^K \pi_{k}^{\,n_{k}}\Bigr)
\times
\Bigl(\prod_{k=1}^K \pi_{k}^{\,\alpha_{k}-1}\Bigr)
\;=\;
\prod_{k=1}^K 
\pi_{k}^{\,n_{k}+\alpha_{k}-1},
\]
i.e.\ a \(\mathrm{Dirichlet}(n_{1}+\alpha_{1},\dots,n_{K}+\alpha_{K})\).  The posterior mean of \(\pi_{k}\) becomes
\[
\pi_{k}^{(\mathrm{mean})}
\;=\;
\frac{n_{k} + \alpha_{k}}{\sum_{j=1}^K \bigl(n_{j} + \alpha_{j}\bigr)}.
\]
Consequently, we adopt $q(\mathbf{y}\mid \tilde{\mathbf{x}})\;=\;\mathrm{Cat}\bigl(\boldsymbol{\pi}^{(\mathrm{mean})}\bigr)$
as our final predictive distribution, which balances empirical label frequencies with the original vPGM's numerical probabilities.

\subsection{Optimizing \(\lambda\) via a Differentiable Calibration Loss}
\label{subsec:optimizing_lambda}
One key limitation of this posterior distribution is its reliance on a manually tuned \(\lambda\), which governs how strongly the vPGM's numerical probabilities influence the final outcome. To automate this process and improve calibration, we introduce a differentiable calibration loss that learns \(\lambda\) through gradient‐based optimization.      

Specifically, we minimize the following loss function with respect to $\lambda$:
\begin{equation}
\label{eq:calibration_loss}
\mathcal{L}\bigl(\boldsymbol{\pi} (\lambda)\bigr)
\;=\;
\mathcal{L}_{c}\bigl(\boldsymbol{\pi}(\lambda)\bigr)
\;+\;
\beta \,\mathcal{L}_{v}\bigl(\boldsymbol{\pi}(\lambda)\bigr),
\end{equation}
where \(\boldsymbol{\pi} (\lambda) = (\pi_1^{(\mathrm{mean})},\dots,\pi_K^{(\mathrm{mean})}) \) is the posterior‐mean vector, \(\mathcal{L}_c\) is a standard classification loss (e.g., cross‐entropy), and \(\mathcal{L}_v\) is a differentiable class‐wise alignment term; $\beta$ is a hyperparameter balancing the two losses. Let \(j\) index the categories, and let \(\bar{\pi}_j = \frac{1}{n}\sum_{i=1}^n \pi_{j}^{(i)}\) be the average predicted probability of class \(j\) over a mini‐batch of size \(n\). Likewise, let \(\bar{y}_j = \frac{1}{n}\sum_{i=1}^n y_{j}^{(i)}\) be the empirical fraction of class \(j\), where \(y_{j}^{(i)}\in\{0,1\}\) indicates whether sample \(i\) belongs to class \(j\). Inspired by class‐wise expected calibration error \cite{kull2019beyond}, which aligns predictions to empirical frequencies on a per‐category basis but whose binning procedure impedes differentiability, we define: 
\begin{equation}
\label{eq:calibration_alignment}
\mathcal{L}_{v}\bigl(\boldsymbol{\pi}\bigr)
\;=\;
\frac{1}{K}
\sum_{j=1}^{K}
\Bigl|
  \bar{\pi}_j
  \;-\;
  \bar{y}_j
\Bigr|,
\end{equation}
using a bin‐free version of class-wise expected calibration error.

To minimize \(\mathcal{L}\bigl(\boldsymbol{\pi}\bigr)\) with respect to \(\lambda\), we employ a quasi‐Newton method (e.g., L-BFGS) \cite{broyden1967quasi}. This second‐order gradient‐based solver converges more rapidly than simple gradient descent.

\begin{theorem}[\textbf{Global Optimum Implies  Perfect ECE}]
\label{thm:ecc_zero}
Let \(\{(\mathbf{u}_i,y_i)\}_{i=1}^{n}\) be the training set with
features \(\mathbf{u}_i\in\mathbb{R}^{d}\) and one–hot labels
\(y_{ik}\). For any parameter vector \(\theta\), let
\(g_{\theta}:\mathbb{R}^{d}\!\to\!\Delta^{K-1}\) be a function that
produces class probabilities
\(\widehat{p}_{ik}(\theta)=g_{\theta}(\mathbf{u}_i)_k\).
The empirical version of \eqnref{eq:calibration_loss} is
\[
\mathcal{L}(\theta)=
-\frac{1}{n}\sum_{i=1}^{n}\sum_{k=1}^{K}
      y_{ik}\,\log\widehat{p}_{ik}(\theta)
+
\beta\,
\frac{1}{K}\sum_{k=1}^{K}
      \bigl|\bar{\widehat{p}}_{k}(\theta)-\bar{y}_{k}\bigr|,
\]
where \( \beta>0, \bar{\widehat{p}}_{k}(\theta)=\tfrac{1}{n}\sum_{i}\widehat{p}_{ik}(\theta)\),
and
\(\bar{y}_{k}=\tfrac{1}{n}\sum_{i}y_{ik}\). Then a parameter vector \(\theta^{\star}\) is a global minimiser of
\(\mathcal{L}\) \emph{iff}
\[
\begin{aligned}
&\widehat{p}_{ik}(\theta^{\star}) = 
\frac{1}{\sum_{i'=1}^{n}
       \mathbb{I}_{\{\mathbf{u}_i = \mathbf{u}_{i'}\}}}
\sum_{i'=1}^{n}
      y_{i'k} \mathbb{I}_{\{\mathbf{u}_i = \mathbf{u}_{i'}\}}
\\
&\text{for every } i\in\{1,\dots,n\},\;k\in\{1,\dots,K\},
\end{aligned}
\]
where \(\mathbb{I}(\mathbf{u}_i = \mathbf{u}_{i'})\) is an indicator function equal to 1 if the feature inputs \(\mathbf{u}_i\) and \(\mathbf{u}_{i'}\) are identical, and 0 otherwise.
In that case, the class-wise expected calibration error  
\(
\operatorname{ECE}_{\mathrm{class}}(\theta)
\triangleq \frac{1}{K}\sum_{k}|\bar{\widehat{p}}_{k}(\theta)-\bar{y}_{k}|
\)
satisfies \(\operatorname{ECE}_{\mathrm{class}}(\theta^{\star})=0\).
\end{theorem}

The proof is provided in Appendix. Although the cross-entropy term in the loss function \eqnref{eq:calibration_loss} pulls predictions toward one-hot labels while the calibration term enforces class-wise average alignment, Theorem \ref{thm:ecc_zero} shows that both objectives can attain their minima simultaneously.

\section{Experiments}
We evaluate the efficacy of the proposed vPGM and BayesVPGM in modeling uncertainty across three agentic reasoning tasks. The first, a closed-ended task named ScienceQA \cite{lu2022learn}, and the second, an open-ended task named ChatCoach \cite{huang2024acl}, both require reasoning with undisclosed information from multiple sources. We then introduce a negative control experiment derived from
A-OKVQA \cite{schwenk2022okvqa} to investigate whether latent variables can enhance confidence calibration by detecting mismatches in
the presence of misinformation. See Appendix for the detailed experimental configurations.

\subsection{Science Question Answering}
The Science Question Answering (ScienceQA) benchmark, introduced by \cite{lu2022learn}, serves as a comprehensive benchmark for multi-modal question answering across a diverse range of scientific disciplines, including physics, mathematics, biology, and the humanities. It features 4,241 question-answer pairs that cover various topics and contexts. This task demands the integration of information from multiple sources or LLM agents (e.g., Bing search results, image captions), a process that can introduce errors and increase the complexity of reasoning. Given these challenges, ScienceQA serves as an ideal testbed for evaluating how effectively vPGM identifies latent structures and model uncertainties. See Appendix for the more detailed experimental setups.

\paragraph{Baseline Methods} We compare vPGM/BayesVPGM with the following baseline methods:
\begin{itemize}[noitemsep, labelwidth=!, topsep=0pt, leftmargin=*]
    \item \textbf{Chain-of-Thought}\: This is one of the non-tool-augmented LLMs: Chain-of-Thought (CoT) prompting \cite{wei2022chain} equipped with verbalized confidence estimation by prompting it to provide a numerical confidence for the selected answer.
    \item \textbf{Chameleon}\: This is based on a tool-augmented LLM: Chameleon \cite{lu2023chameleon}, and we equip it with verbalized confidence estimation.
    \item \textbf{Chameleon+}\: It extends Chameleon with a state-of-art uncertainty quantification framework based on the combination of verbalized confidence estimation and self-consistency measurement \cite{wang2022self}, as recommended in \cite{xiong2023llms}.
\end{itemize}

\paragraph{Evaluation Metrics} In line with previous evaluation settings in \cite{naeini2015obtaining,guo2017calibration,xiong2023llms} on confidence calibration, we adopt the expected calibration error (\textbf{ECE}) to evaluate model confidence, represented as numeric probabilistic predictions. The ECE quantifies the divergence between the predicted probabilities and the observed accuracy across each confidence levels (bins). Throughout our experiments, we fix the number of confidence bins as 10 with uniform confidence contribution across bins. In addition, we evaluate the capability of a given method in solving problems correctly by measuring the accuracy (\textbf{Acc.}).

\begin{table}[htb!]
    \tabcolsep=0.13cm
    \centering
    \begin{tabular}{l  c  c  c  c}
        \toprule
        Method               & \(N\) & \(M\) & Acc. \(\uparrow\) & ECE \(\downarrow\) \\
        \midrule
        CoT                & –     & 1     & 84.63                   &  8.96                               \\
        Chameleon          & –     & 1     & 85.29                   &  9.62                               \\
        Chameleon+         & –     & 3     & 85.17                   &  8.65                               \\
        vPGM (Ours)              & 2     & 3     & 85.49                   &  2.31                               \\
        vPGM (Ours)              & 3     & 3     & \underline{86.38}       &  1.67                               \\
        vPGM (Ours)              & 4     & 3     & \textbf{86.54}          &  2.15                               \\
        BayesVPGM (Ours)          & 2     & 3     & 85.49                   &  1.81                               \\
        BayesVPGM (Ours)         & 3     & 3     & 86.38                   & \textbf{1.05}                    \\
        BayesVPGM (Ours)         & 4     & 3     & \textbf{86.54}          & \underline{1.50}                       \\
        \bottomrule
    \end{tabular}
    \caption{Accuracy ($\%$) and ECE $(\times10^2)$ on ScienceQA for different methods and numbers of latent variables \(N\).  \(M\) is the number of sampled responses. The best and second-best results within each base model are \textbf{bolded} and \underline{underlined}, respectively. Llama3-8B-Instruct \cite{dubey2024llama} serves as our test-time engine. See appendix for results using other LLMs}
    \label{tbl-sciqa-combined}
\end{table}

\begin{figure}[ht]
    \begin{center}
        \resizebox{0.35\textwidth}{!}
        {\includegraphics{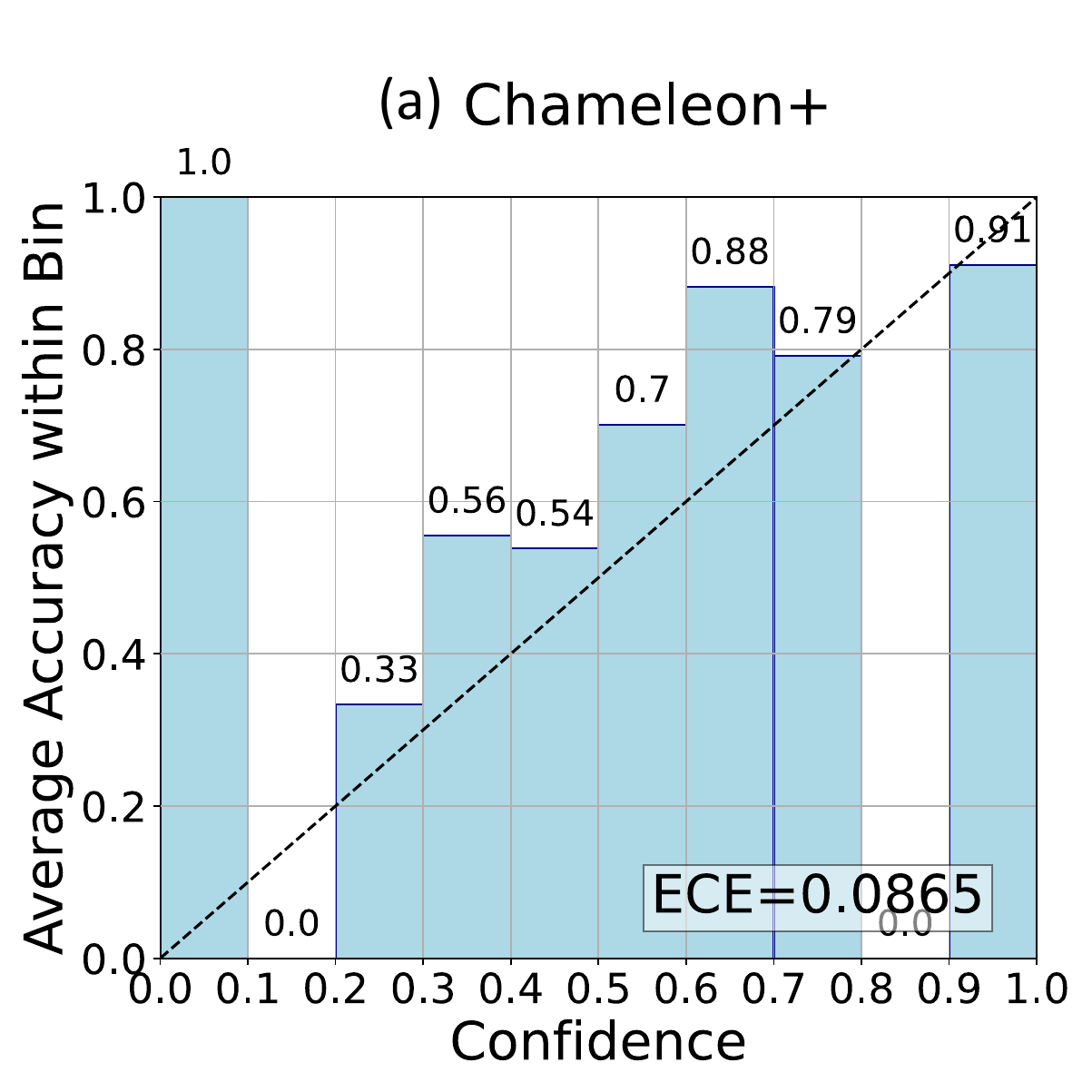}}
        \resizebox{0.35\textwidth}{!}
        {\includegraphics{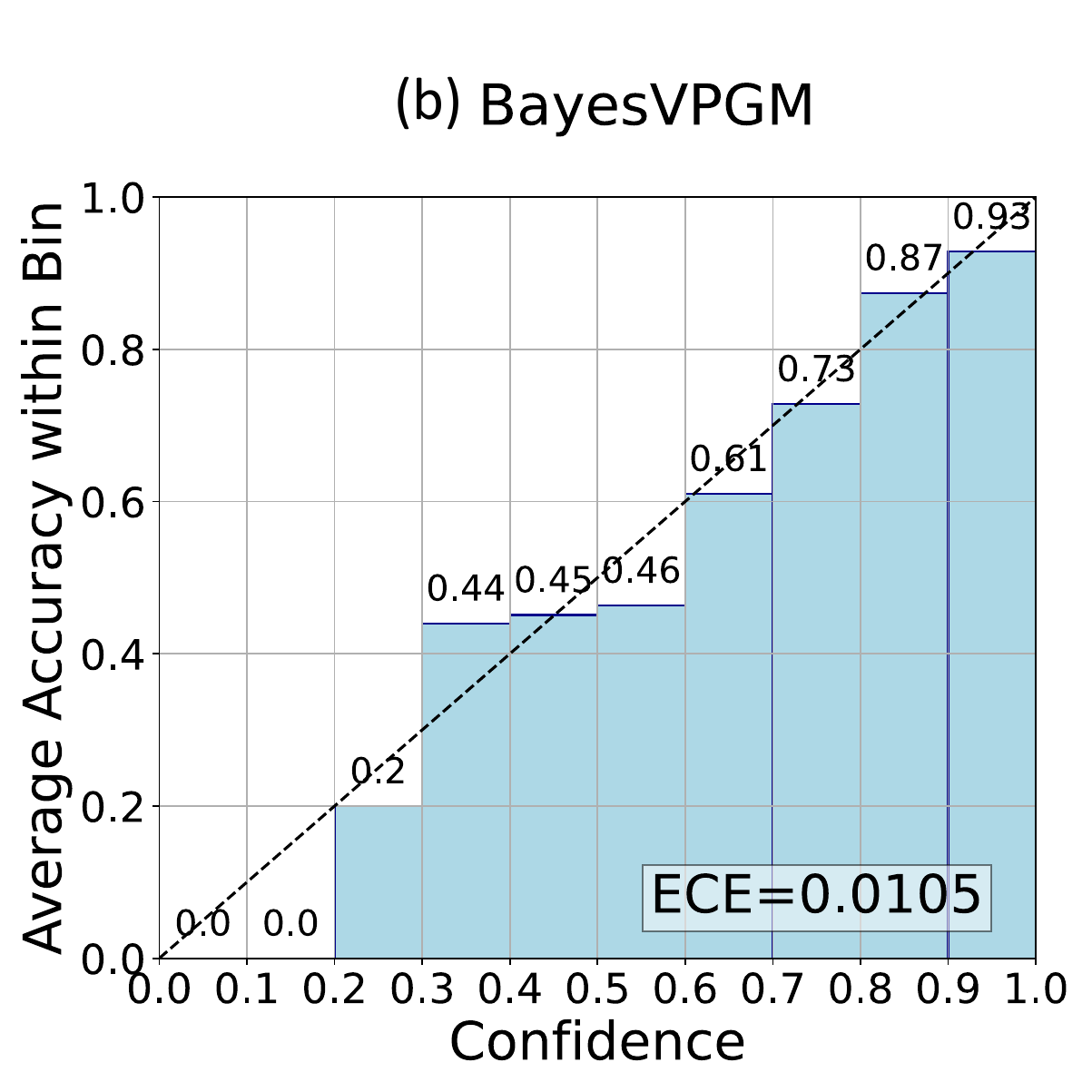}}
        \caption{Reliability diagrams of (a) Chameleon+ and (b) BayesVPGM ($N=3,M=3$) on ScienceQA (see the appendix for diagrams of Chameleon and vPGM). BayesVPGM achieve a much lower ECE comparing to Chameleon+ and approaches to the ideal confidence calibration curve (the diagonal dashed line).}
        \label{fig:sciqa3568_ece}
    \end{center}
\end{figure}

\paragraph{Results} \tabref{tbl-sciqa-combined} details the performance of different methods on the ScienceQA dataset. It shows that Chameleon results in the highest (worst) ECE ($\times 10^2$) of 9.62, indicating serious overconfidence issues in handling complex reasoning tasks, even with the assistance of external tools. In comparison, our vPGM outperforms these methods in both accuracy and ECE, due to its superior ability to capture latent structural information that other baseline methods overlook. Figure \ref{fig:sciqa3568_ece} shows the reliability diagram for vPGM and BayesVPGM, demonstrating its near-perfect alignment with the ideal calibration curve across all bins, highlighting its precision in confidence calibration (see the Appendix for the ablation results and the token-level computational costs).

\paragraph{Qualitative Study on the Inferred Latent Variables} Figure \ref{fig:overview_inference} shows a case study of BayesVPGM's inference capabilities to qualitatively assess the model's ability to utilize latent structural information for improving confidence estimation. Here vPGM employs its latent variables to critically assess the relevance of retrieved information. For example, when faced with irrelevant data from external tools such as Bing search or inaccurate captions from image captioners, the baseline, Chameleon, erroneously maintains high confidence in its predictions. In contrast, BayesVPGM carefully adjusts its confidence, assigning lower probabilities when essential contextual knowledge is missing or incorrect, a process that is particularly effective through the inference of latent variables $Z_1$ and $Z_2$. These observations highlight the significance of inferring latent structures to improve the reliability of compositional reasoning systems.

\begin{table*}[ht]
    \centering
    \tabcolsep=0.14cm
    \begin{tabular}{lcccccc}
    \toprule
    \multirow{2}{*}{Method} & \multicolumn{3}{c}{Detection} & \multicolumn{3}{c}{Correction} \\
     & BLEU-2 & Rouge-L & BERTScore & BLEU-2 & Rouge-L & BERTScore \\
    \midrule
    Instruction Prompting & 27.4 & {3.3} & 67.6 & 1.4 & 2.1 & {61.6}\\
    Vanilla CoT & 17.7 & 2.7 & 64.1 & 0.1 & \textbf{2.3} & 58.1 \\
    Zero-shot CoT & 27.6 & 1.9 & 69.0 & \textbf{3.0} & 0.9 & 58.8 \\
    GCoT & 34.2 & \textbf{3.7} & 72.4 & 1.6 & {2.0} & 65.4\\
    vPGM (Ours) & \textbf{37.2} & 2.3 & \textbf{76.3} & 1.7 & 2.0 & \textbf{68.3}\\
    \midrule
    Human & 76.6 & 6.0 & 90.5 & 33.5 & 3.6 & 84.1\\
    \bottomrule
    \end{tabular}
    \caption{Results of various methods on the detection and correction of medical terminology errors. }
    \label{tab:bcot_result}
\end{table*}

\subsection{Communicative Medical Coaching}
The Communicative Medical Coaching benchmark, ChatCoach, introduced in \cite{huang2024acl}, establishes a complex multi-agent dialogue scenario involving doctors, patients, and a medical coach across 3,500 conversation turns. The medical coach is tasked with detecting inaccuracies in medical terminology used by doctors (\textbf{detection task}) and suggesting appropriate corrections (\textbf{correction task}). These tasks require integrating external medical knowledge, inherently introducing uncertainty into response formulation. This benchmark was chosen to test vPGM's ability to generalize across complex open-ended reasoning tasks. BayesVPGM is not applied in this setting, as such a model assumes the output to be a categorical distribution. See the Appendix for more details on experiments and implementation.

\paragraph{Baseline Methods} For comparative analysis, we benchmark vPGM against these approaches: 
\begin{itemize}[noitemsep, labelwidth=!, topsep=0pt, leftmargin=*]
    \item \textbf{Vanilla Instruction Prompting:} This method involves prompting the LLM with direct instructions for dialogue generation.
    \item \textbf{Zero-shot Chain of Thought (CoT) \cite{kojima2022large}:} A straightforward CoT approach where the LLM is prompted to sequentially articulate a reasoning chain.
    \item \textbf{Vanilla CoT \cite{wei2022chain}:} This method builds upon the basic CoT by providing the LLM with a set of examples that include detailed reasoning steps.
    \item \textbf{Generalized CoT (GCoT) \cite{huang2024acl}:} An advanced version of CoT, designed to improve the generation of structured feedback and integration of external knowledge effectively. It represents a state-of-the-art method in the ChatCoach benchmark.
\end{itemize}

\paragraph{Evaluation Metrics} We follow \cite{huang2024acl} to employ conventional automated metrics \textbf{BLEU-2}, \textbf{ROUGE-L}, and \textbf{BERTScore}. BLEU-2 is employed to measure the precision of bi-gram overlaps, offering insights into the lexical accuracy of the generated text against reference answers. ROUGE-L is used to assess sentence-level similarity, focusing on the longest common subsequence to evaluate structural coherence and the alignment of sequential n-grams. Additionally, BERTScore is applied for a semantic similarity assessment, utilizing BERT embeddings to compare the generated outputs and reference texts on a deeper semantic level.  As specified in \cite{huang2024acl}, we use GPT-4 to extract medical terminology errors and corresponding corrections in the feedback from Coach Agents. Automated metrics are then calculated based on these extracted elements in comparison to human annotations. 

\paragraph{Results} We present the performance of various methods in Table~\ref{tab:bcot_result}. The noticeable difference between machine-generated outputs and human benchmarks across all metrics highlights the inherent challenges in communicative medical coaching. In the detection of medical terminology errors, vPGM leads with superior BLEU-2 (37.2) and BERTScore (76.3), underscoring its proficiency in identifying inaccuracies. In the correction task, while vPGM achieves a standout BERTScore of 68.3, surpassing all baselines, it scores lower on BLEU-2 and ROUGE-L. This variation is attributed to the ambiguity in doctors’ inputs, which can yield multiple valid responses, affecting metrics that rely on exact matches.

\subsection{A-OKVQA Negative Control: Studying Latent Variables Under Misinformation}
\paragraph{Data Simulation} A-OKVQA \cite{schwenk2022okvqa} is a Visual Question Answering dataset that challenges models to perform commonsense reasoning about a scene, often beyond the reach of simple knowledge-base queries. Crucially, it provides ground-truth image captions and rationales for each question. We leverage these annotations to construct a negative control experiment: \textbf{A-OKVQA-clean} (603 data points) retains the correct image caption and rationale (near single-hop reasoning), while 
\textbf{A-OKVQA-noisy} (603 data points) randomly shuffles the rationale, thus introducing misinformation and forcing a multi-hop check for consistency. In this experiment, we adopt a vPGM with 2 latent variables (see the Appendix for the inference prompt and an example query). Refer to the Appendix for more details on data configurations.

\paragraph{Overall Performance Under Noisy Conditions}
Table~\ref{tab:aokvqa_gp} shows the overall accuracy (\textbf{Acc.}) 
and expected calibration error (\textbf{ECE}) on the A-OKVQA-noisy dataset. 
Both vPGM and BayesVPGM outperform Chameleon+ on accuracy (61.03\% vs.\ 59.04\%) and yield lower ECE, indicating that latent variables detect mismatch and improve confidence calibration.

\paragraph{Mismatch Detection Through \boldmath{$Z_2$}} To investigate how latent variables facilitate mismatch detection, we track \(P\bigl(Z_2 \mid \mathrm{Pa}(Z_2)\bigr)\), where \(Z_2\) indicates whether the rationale is aligned with the image caption. As shown in Table~\ref{tab:aokvqa_lv}, the mean probability of \(Z_2\) is considerably higher in the \emph{Clean} set than in the \emph{Noisy} set (0.86 vs.\ 0.42), and mismatch identification accuracy in the \emph{Noisy} condition reaches 87\%. These findings demonstrate BayesVPGM’s capacity to robustly detect cases with inconsistencies or irrelevant content.

\paragraph{Latent Variable Correlation Analysis} We additionally compute Pearson correlations (Pcc.) between numerical conditional probabilities of the latent variables ($Z_1$ and $Z_2$) and the final answer \(\mathbf{Y}\). In the \emph{Noisy} case, \(\text{Pcc}(Z_2,\mathbf{Y})\)  surpasses \(\text{Pcc}(Z_1,\mathbf{Y})\) (0.55 versus 0.35), indicating that \(Z_2\) exerts a stronger influence on the final prediction when mismatches are present. Conversely, in the \emph{Clean} subset, \(Z_1\) and \(Z_2\) exhibit nearly equal correlation with \(\mathbf{Y}\), yet about 22\% of the \emph{Clean} data is incorrectly flagged by \(Z_2\) as mismatched, potentially introducing noisy confidence adjustments at \(\mathbf{Y}\). This suggests a trade-off: while latent variables excel at detecting misinformation and improving calibration in \emph{Noisy} settings, they can slightly degrade calibration when no mismatch actually exists.

\begin{table}[ht]
    \centering
    \begin{tabular}{lcc}
        \toprule
        Method & Acc. & ECE \\
        \midrule
        Chameleon+ & 59.04 & 11.75 \\
        vPGM (Ours) & \textbf{61.03} & 10.54 \\
        BayesVPGM (Ours) & \textbf{61.03} & \textbf{9.85} \\
        \bottomrule
    \end{tabular}
    \caption{General Performance on A-OKVQA-noisy data (accuracy in $\%$ and ECE in $\times10^2$).}
    \label{tab:aokvqa_gp}
\end{table}
\begin{table}[ht]
    \centering
    \begin{tabular}{lcc}
    \toprule
                                & Clean & Noisy \\ \midrule
    Mean \(P\bigl(Z_2 \mid \mathrm{Pa}(Z_2)\bigr)\) & 0.86  & 0.42  \\
    Noise Identification Acc.   & 78\% & 87\% \\
    \(\text{Pcc}\bigl(Z_1,\mathbf{Y}\bigr)\)            & 0.50  & 0.35  \\
    \(\text{Pcc}\bigl(Z_2,\mathbf{Y}\bigr)\)            & 0.51  & 0.55  \\
    \bottomrule
    \end{tabular}
    \caption{Analysis of the latent variables on A-OKVQA-clean and A-OKVQA-noisy.}
    \label{tab:aokvqa_lv}
\end{table}

\section{Conclusion}
We introduce verbalized Probabilistic Graphical Model (vPGM), a Bayesian agentic framework that (1) directs LLM agents to simulate core principles of Probabilistic Graphical Models (PGMs) through natural language and (2) refines the resulting posterior distributions via numerical Bayesian inference. Applied within agentic workflows, vPGM enables LLM agents to perform probabilistic latent-variable reasoning with calibrated uncertainty. This approach discovers latent variables and dependencies without requiring extensive domain expertise, making it well-suited to settings with limited assumptions. Our empirical results on agentic reasoning tasks demonstrate substantial improvements in terms of both confidence calibration and text generation quality. These results highlight the potential of merging Bayesian principles with LLM agents to enhance AI systems' capacity for modeling uncertainty and reasoning under uncertainty.

\section*{Acknowledgments}
We thank all reviewers, SPC, and AC for their valuable comments. S.B. acknowledges funding from the MRC Centre for Global Infectious Disease Analysis (reference MR/X020258/1), funded by the UK Medical Research Council (MRC). This UK funded award is carried out in the frame of the Global Health EDCTP3 Joint Undertaking. S.B. is funded by the National Institute for Health and Care Research (NIHR) Health Protection Research Unit in Modelling and Health Economics, a partnership between UK Health Security Agency, Imperial College London and LSHTM (grant code NIHR200908). H.W. is partially supported by Amazon Faculty Research Award, Microsoft AI \& Society Fellowship, NSF CAREER Award IIS-2340125, NIH grant R01CA297832, and NSF grant IIS-2127918. We acknowledge support from OpenAI’s Researcher Access Program. Disclaimer: ``The views expressed are those of the author(s) and not necessarily those of the NIHR, UK Health Security Agency or the Department of Health and Social Care.'' S.B. acknowledges support from the Novo Nordisk Foundation via The Novo Nordisk Young Investigator Award (NNF20OC0059309). S.B. acknowledges the Danish National Research Foundation (DNRF160) through the chair grant. S.B. acknowledges support from The Eric and Wendy Schmidt Fund For Strategic Innovation via the Schmidt Polymath Award (G-22-63345) which also supports H.H. and L.M.

\bibliography{main}


\end{document}